%% file: Formatting-Instructions-LaTeX-2026.tex
\documentclass[letterpaper]{article} 
\usepackage{aaai2026}  
\usepackage{times}  
\usepackage{helvet}  
\usepackage{courier}  
\usepackage[hyphens]{url}  
\usepackage{graphicx} 
\usepackage{natbib}  
\usepackage{caption} 
\usepackage{algorithm}
\usepackage{algorithmic}

\usepackage{algorithm}
\usepackage{algorithmic}
\usepackage{amsmath}

\usepackage{soul}
\usepackage[utf8]{inputenc}
\usepackage[T1]{fontenc}
\usepackage{xcolor}
\usepackage{booktabs}
\usepackage{multirow}
\usepackage{xparse,mleftright}
\usepackage{amssymb}
\usepackage{marvosym}

\usepackage{color}
\definecolor{BurntOrange}{RGB}{0.8, 0.33, 0}
\definecolor{OliveGreen}{RGB}{85, 107, 47}
\definecolor{Plum}{RGB}{139, 102, 139}
\definecolor{Blue}{RGB}{0, 0, 255}

\usepackage{newfloat}
\usepackage{listings}
\DeclareCaptionStyle{ruled}{labelfont=normalfont,labelsep=colon,strut=off} 
\floatstyle{ruled}
\newfloat{listing}{tb}{lst}{}
\floatname{listing}{Listing}
\title{InteChar: A Unified Oracle Bone Character List \\ for Ancient Chinese Language Modeling}
\author{
    Xiaolei Diao\textsuperscript{\rm 1, \rm 2},
    Zhihan Zhou\textsuperscript{\rm 2},
    Lida Shi\textsuperscript{\rm 2}, 
    Ting Wang\textsuperscript{\rm 3}, 
    Ruihua Qi\textsuperscript{\rm 2}, 
    Hao Xu\textsuperscript{\rm 2}, 
    Daqian Shi\textsuperscript{\rm 1}\thanks{Corrosponding Author.}\\
}
\affiliations{
    \textsuperscript{\rm 1}Queen Mary University of London \quad\quad
    \textsuperscript{\rm 2}Jilin University \quad\quad
    \textsuperscript{\rm 3}Tongji University \quad\quad

}

\usepackage{bibentry}

\begin{document}

\maketitle

\begin{abstract}
Constructing historical language models (LMs) plays a crucial role in aiding archaeological provenance studies and understanding ancient cultures. However, existing resources present major challenges for training effective LMs on historical texts. First, the scarcity of historical language samples renders unsupervised learning approaches based on large text corpora highly inefficient, hindering effective pre-training. Moreover, due to the considerable temporal gap and complex evolution of ancient scripts, the absence of comprehensive character encoding schemes limits the digitization and computational processing of ancient texts, particularly in early Chinese writing. To address these challenges, we introduce InteChar, a unified and extensible character list that integrates unencoded oracle bone characters with traditional and modern Chinese. InteChar enables consistent digitization and representation of historical texts, providing a foundation for robust modeling of ancient scripts. To evaluate the effectiveness of InteChar, we construct the Oracle Corpus Set (OracleCS), an ancient Chinese corpus that combines expert-annotated samples with LLM-assisted data augmentation, centered on Chinese oracle bone inscriptions. Extensive experiments show that models trained with InteChar on OracleCS achieve substantial improvements across various historical language understanding tasks, confirming the effectiveness of our approach and establishing a solid foundation for future research in ancient Chinese NLP.
\end{abstract}


\section{Introduction}
\label{sec:Introduction}
\input{sections/s1-introduction}

\section{Related Work}
\label{sec:Related Work}
\input{sections/s2-related_work}


\section{The Proposed Method}
\label{sec:Method}
\input{sections/s4-method}

\section{Experiments and Discussions}
\label{sec:Experiments}
\input{sections/s5-experiments}
\section{Conclusion}
\label{sec:Conclusion}
\input{sections/s6-conclusion}
\bigskip

\bibliography{aaai2026}

\end{document}

%% file: sections/s1-introduction.tex
Ancient script research has long served as a cornerstone for cultural heritage preservation and the advancement of historical linguistics, enabling scholars to decode lost histories and gain insight into ancient cultures through the inscriptions found on archaeological artifacts. Traditional approaches to this research are largely influenced by human cognitive and learning limitations, which restrict the efficiency of data processing and significantly hinder the decipherment of unknown characters \cite{diao2023rzcr}. In contrast, recent advances in natural language understanding have demonstrated significant advantages in processing vast and complex corpora, motivating researchers to apply these techniques to large-scale ancient text processing tasks \cite{tian2021anchibert,  stopponi2024natural}. One of the key research directions in this field is the development of historical language models specifically designed to comprehend ancient textual materials, thereby facilitating tasks such as archaeological inference, historical reconstruction, and cultural analysis \cite{ross2023new, koc2025exploring}.

\begin{figure}[!t]
\centering
\includegraphics[width=1\linewidth]{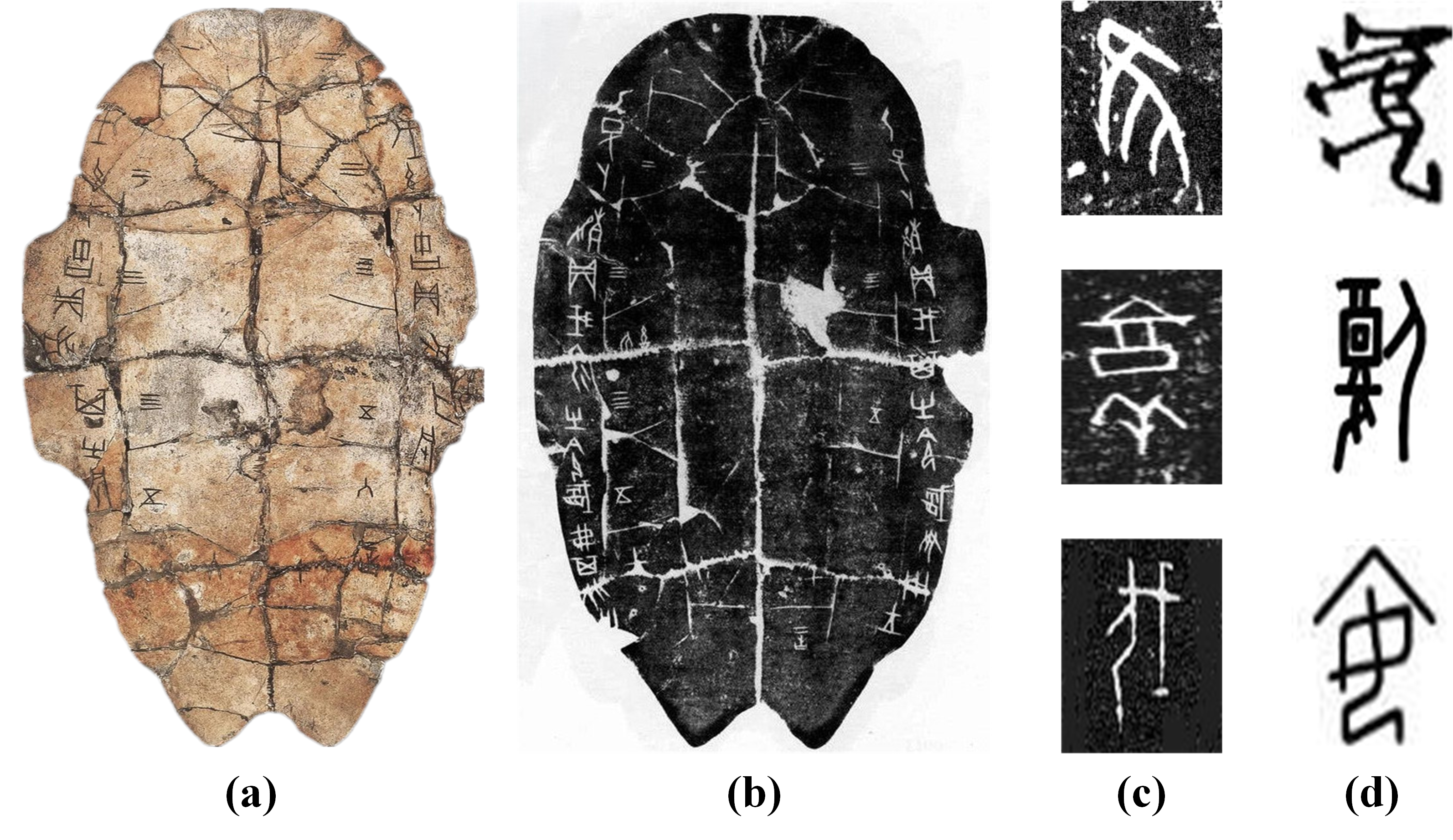}
\caption{Examples of various types of OBI images. (a) An entire piece of oracle bone. (b) An entire piece of oracle bone rubbing. (c) Oracle character rubbing images. (d) Handwritten oracle character images.}
\vspace{-0.3cm}
\label{fig:1}
\end{figure}

Collecting and constructing appropriate corpora is fundamental to the development of effective historical language models. However, due to the great antiquity of ancient scripts and their complex evolution over time, excavated documents represented by Oracle Bone Inscriptions (OBI) frequently contain a large number of unencoded characters, which poses substantial challenges for digitization. As illustrated in Figure~\ref{fig:1}, the common approach currently involves storing these characters as images \cite{shi2022rcrn, guan2024deciphering, wang2024open, gao2024linking}, which are often in handwritten or rubbing form, rather than as machine-encoded text. Furthermore, the limited number and preservation issues of excavated artifacts result in relatively scarce corpus samples for these ancient languages \cite{chi2022zinet, diao2023toward}. For example, only about 5,000 complete oracle bone pieces have been unearthed, yielding merely 15,000 sentences that contain more than five characters. Consequently, the combination of vast numbers of unencoded characters and the scarcity of sentence-level samples poses a significant challenge to the construction of an effective and usable corpus for ancient texts.

Research on ancient language models remains at an early stage. Some studies \cite{guo2023towards, wang2023gujibert, liu2024sikugpt} have attempted to repurpose modern character encoding schemes to represent ancient scripts, training word embeddings and constructing corpora based on these representations. However, such approaches typically cover only a subset of high-frequency ancient characters, while excluding a large number of low-frequency or unencoded characters. This exclusion is not trivial: in the context of ancient texts, where language resources are extremely limited and the writing system often highly contextual, even a single low-frequency character may carry unique semantic, historical, or cultural significance \cite{diao2023rzcr}. Each character, regardless of its frequency, can be crucial for accurately reconstructing meaning, understanding rare expressions, or tracing cultural and linguistic developments. Therefore, models that omit these characters suffer from incomplete representations and risk significant information loss during training \cite{zhang2023can}. Other efforts have explored directly training language models on limited transcriptions from excavated artifacts \cite{chi2022zinet}, but these are similarly constrained by the scarcity of annotated data, which makes unsupervised pre-training inefficient and affects effective semantic learning. The extremely low frequency of many ancient characters further complicates this challenge, as their meanings cannot be reliably inferred from the surrounding context alone.

To address the above challenges, we propose a novel corpus to support historical language modeling. A key component of our approach is InteChar, a unified and extensible character set that incorporates oracle bone characters not covered by existing encoding standards. We introduce a standardized digitization pipeline that converts scanned images containing OBIs into machine-readable text, encoded in a format compatible with modern character standards. InteChar enables a consistent and comprehensive digital representation of ancient scripts, providing a solid foundation for subsequent corpus construction and model training. Furthermore, although many ancient characters appear infrequently, their latent mappings to modern Chinese can be reinforced using data distillation techniques within a pre-training paradigm. By augmenting the corpus with enhanced samples and integrating specialized pre-training strategies, models can acquire semantic representations of these rare characters more rapidly. Following this strategy, we construct OracleCS, a corpus of excavated transcriptions of oracle bone inscriptions designed to support the training of historical Chinese language models in low-resource settings. In addition, we construct a multi-task benchmark to quantitatively evaluate the performance of language models on understanding ancient scripts. To our knowledge, this work is the first to systematically incorporate excavated oracle characters, including undeciphered ones, into LM‘s evaluation pipelines. The main contributions include:
\begin{itemize}
    \item We construct InteChar, a unified and extensible Unicode-compatible character list that integrates previously unencoded oracle bone characters alongside traditional and modern Chinese characters, enabling consistent and comprehensive digitization of ancient texts. 

    \item  Based on InteChar, we build OracleCS, a corpus that prominently features oracle bone transcriptions, and develop a benchmark to systematically evaluate language models on ancient Chinese language understanding.

    \item  Extensive experiments demonstrate that models trained on OracleCS with InteChar significantly outperform baselines across both embedding-based evaluations and downstream fine-tuning tasks.

\end{itemize}

%% file: sections/s2-related_work.tex

\subsection{Ancient Character Tasks}

In recent years, the study of ancient Chinese characters has gained increasing attention due to its value in historical linguistics, archaeology, and cultural heritage. Advances in deep learning have greatly supported this field, especially in tasks like character recognition \cite{lin2022radical}, detection \cite{yue2025ancient}, and restoration \cite{shi2022charformer, li2023towards}. Early studies primarily focused on image-based recognition tasks, such as oracle bone inscription classification using computer vision techniques. For example, Zhang et al. \cite{zhang2021ai} employed a Siamese network to match rubbing images with template images from oracle character databases. Later studies have expanded beyond recognition to include glyph identification and structure analysis. RZCR \cite{diao2023rzcr} combines radical and structure features for character recognition, while LUC \cite{gao2024linking} added radical and domain-specific features to improve character retrieval. \cite{chi2022zinet} propose an ancient Chinese knowledge graph ZiNet that links glyphs and radicals across time. Generative methods have also been explored. \cite{guan2024deciphering} proposed a diffusion-based approach to generate possible modern forms from ancient characters, helping with transcription and understanding how characters evolved. Despite these advances, such studies only focus on deciphered characters. The interpretation of entirely unknown characters remains an open and challenging problem.

\subsection{Ancient Chinese Corpus Collection}
A key challenge in using NLP for ancient languages is the lack of annotated data. Ancient Chinese texts, e.g., oracle bones and bronze inscriptions, are typically available only as noisy, fragmented images with limited annotations \cite{shi2022rcrn, diao2025oracle}. This low-resource scenario necessitates innovative data collection and augmentation strategies. Recent work on word segmentation has explored using language models to create synthetic training data. For example, \cite{shen2022data} used an LSTM model \cite{hochreiter1997long} to generate labeled samples, and \cite{feng2023ancient} applied distant supervision with parallel texts to build augmented datasets. Expert-annotated corpora are also essential. For instance, the CHisIEC dataset \cite{tang2024chisiec} combines expert knowledge with text analysis to extract relations from historical texts.

\subsection{Benchmarks of Ancient Language Models}
Evaluating the performance of language models trained on ancient scripts poses unique challenges. Recent research has focused on building benchmark datasets to assess model capabilities on ancient Chinese texts. For example, FSPC \cite{FSPC} and CCMP \cite{li2021ccpm} provide corpora for classical poetry comprehension tasks, while CUGE\footnote{\url{https://cuge.baai.ac.cn}.}  \cite{yao2021cuge} extends CCMP by adding a poetry matching subtask. Zinin and Xu \cite{zinin-xu-2020-corpus} compiled historical travel texts and other ancient corpora to enrich data diversity for downstream tasks. Other studies have introduced tasks such as syntactic analysis, topic mining, and sentiment classification \cite{pan2022ancient, wang2022uncertainty, Liu2022Contrastive}.
AC-EVAL \cite{wei2024ac} integrates multiple datasets and tasks into a unified evaluation suite for ancient Chinese understanding. WenMind \cite{cao2024wenmind} focuses on Chinese classical literature and language arts, containing 4,875 question–answer pairs across 42 fine-grained tasks, offering a comprehensive benchmark for evaluating LLMs in this domain. Fùxì \cite{zhao2025f} introduces a benchmark covering 21 tasks aimed at both understanding and generation, including novel tasks such as poetry composition and couplet completion, making it particularly suited for generation-oriented ancient Chinese tasks.

In summary, although significant progress has been made in developing models for the recognition and interpretation of ancient Chinese characters, key challenges remain. The most crucial problem is that all existing benchmarks and corpora only cover encoded texts, ignoring characters from unearthed artifacts that have not been standardized. This greatly limits the ability of language models to learn from ancient texts and manuscripts that contain rich historical and cultural information.

%% file: sections/s4-method.tex
This section introduces the construction of two key resources: the InteChar Unicode character list and the OracleCS corpus. InteChar provides a unified encoding for ancient characters, while OracleCS offers a pretraining dataset for oracle bone script. Both of them form the foundation for downstream modeling and evaluation.


\begin{figure*}[!t]
\centering
\includegraphics[width=1\linewidth]{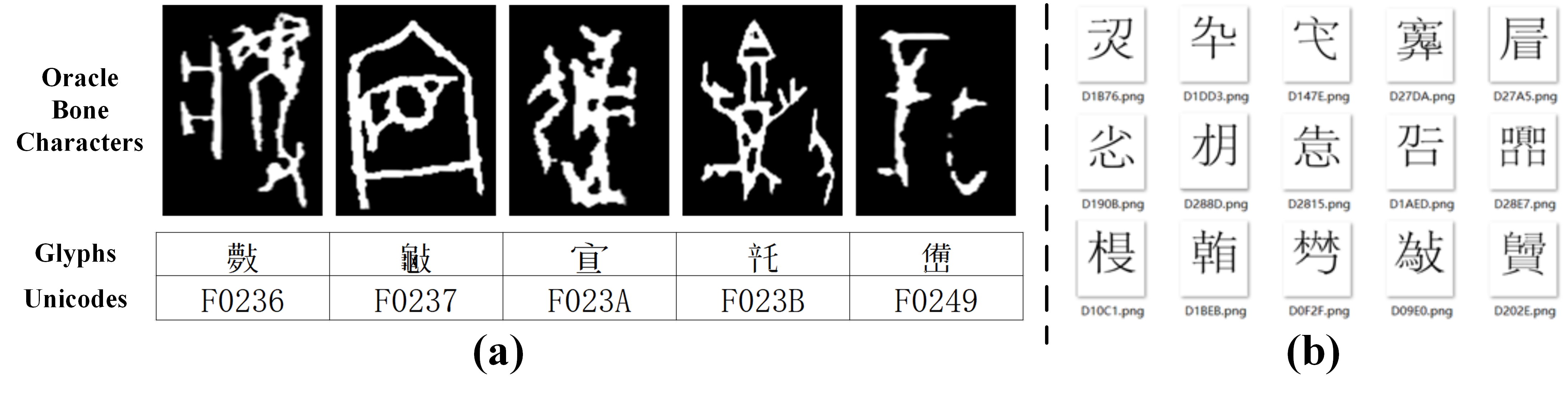}
\caption{Examples of Unicode character lists. (a). Oracle bone Character images with their corresponding standardized glyphs and Unicode in InteChar. (b). Examples of TrueType Font in InteChar.}
\label{fig:InteChar}
\end{figure*}

\subsection{InteChar Character list Construction}
Research on ancient scripts underscores the importance of building a comprehensive Unicode character list to support the development of historical language models. In this work, we construct a unified and structured character set named Integrated Characters (InteChar), which includes oracle bone characters, traditional Chinese characters, and modern simplified characters. InteChar is specifically designed to meet the needs of training models on ancient Chinese texts. We integrate multiple data sources, including modern Unicode character sets, scanned images of oracle bone inscriptions, and specialized font libraries, to produce a complete and standardized character list.

The construction of InteChar follows a four-stage workflow aimed at building a unified and extensible character inventory for ancient Chinese scripts. First, we initialize the character list by loading the official Unicode character set\footnote{\url{https://en.wikipedia.org/wiki/Unicode}}, which serves as the foundational layer for compatibility with modern natural language processing systems. Second, we enrich the list by incorporating encoded characters from widely adopted machine-readable ancient Chinese resources, selecting only those characters that also appear in our curated corpus to ensure relevance and avoid redundancy. Third, we construct entirely new characters for glyphs present in the corpus but absent from existing standards. Finally, we conduct expert-guided proofreading and de-duplication to ensure the integrity, accuracy, and uniqueness of each character entry. The resulting InteChar character set contains both standardized and newly encoded characters, offering robust support for historical language modeling and digital processing of ancient texts.


\noindent \textbf{Initial Character List Construction.}
The initial character list is constructed by loading the official Unicode character set, which defines standardized code points for characters from virtually all major writing systems worldwide. Among them, Unicode includes more than 90,000 encoded CJK characters\footnote{\url{https://en.wikipedia.org/wiki/CJK_Unified_Ideographs}}. This serves as the foundational encoding standard for modern natural language processing, ensuring consistency across platforms and compatibility with mainstream language models.

\noindent \textbf{Integration of Existing Encoded Characters.}
To fully leverage existing resources and reduce manual overhead, we integrate previously encoded ancient characters from widely adopted machine-readable libraries. A primary source is the Zhongjian Library collection published by Zhonghua Book Company\footnote{\url{https://www.ancientbooks.cn/helpcore?font}}, a commonly used resource among paleographers that includes 16 historical font sets initially. These fonts include well-attested glyphs from oracle bones, bronze inscriptions, bamboo manuscripts, and other early Chinese scripts. Our focus is on the oracle bone subset of the Zhongjian Library. To ensure relevance and avoid redundancy, we apply a strict filtering strategy: only characters that appear both in the Zhongjian Library and in our curated corpus are retained. This ensures that all included characters are not only well-formed but also actively used in authentic historical contexts. When a character appears in multiple font sets, we preserve only one representative form to avoid duplication. For each retained glyph, we extract its graphical representation and record the associated metadata in InteChar, including font source and an internally assigned code point within Zhongjian Library. This integration stage enables the reuse of trusted typographic resources and provides a cost-effective foundation for character set expansion.

\noindent \textbf{Construction of New Characters.}
While the integration of existing resources significantly reduces manual effort, a large number of characters in our corpus remain unencoded in both Unicode and historical font libraries, especially from excavated oracle inscriptions. These characters often correspond to undeciphered or low-frequency glyphs that nonetheless carry important contextual and linguistic value. To address this gap, we construct entirely new characters using a semi-automated pipeline that combines computer vision techniques with expert validation. Following previous studies, we rely on three key sources to identify candidate characters for construction: oracle bone images, domain-specific corpora, and existing font resources. Characters that appear in our training corpus but lack a corresponding encoding in Unicode or the Zhongjian Library are flagged as candidates for reconstruction.

We observe that while many oracle glyphs are complex and unique, a large number of their subcomponents, namely radicals, have been extensively studied and can be linked to known components in traditional Chinese characters. Instead of requiring experts to manually trace entire glyphs stroke by stroke, we adopt a radical-based recognition method that detects familiar structural units to enable compositional reconstruction. This approach significantly improves both the efficiency and scalability of new character creation. To efficiently construct new character entries without requiring experts to manually trace every stroke, we apply a radical recognition method that automatically identifies radicals within complex glyphs. By detecting known radicals, we can reconstruct new characters radical by radical, rather than stroke by stroke, significantly accelerating the expansion of the character list. In our implementation, we adopt the radical recognition method proposed by \cite{diao2023toward} to output candidate radicals, which is based on an object detection model trained to identify oracle radicals within noisy, real-world character images. 
The overall pipeline for new character construction consists of the following seven steps: 
\begin{itemize}
    \item Image Collection and Preprocessing: Oracle bone inscription images are collected from archaeological publications and processed via resizing, contrast normalization, and geometric alignment (e.g., vertical flipping).

    \item Radical Recognition: The preprocessed images are passed through the radical recognition model to predict the categories of radicals within each glyph.

    \item Standardization of Components: Mapping predicted radicals to modern Chinese equivalents where applicable, forming an intermediate compositional glyphs.

    \item Expert Verification: Paleographers manually verify the radical composition, correct misclassifications, and make final glyph adjustments based on domain knowledge.

    \item Vectorization: Validated glyphs are redrawn as scalable vector graphics, conforming to a consistent visual style aligned with InteChar’s typographic standard.

    \item Code Point Assignment: Each reconstructed glyph is assigned a new internal code point using a Unicode-style format that supports future interoperability and systematic expansion.

    \item Character Integration: The finalized character is incorporated into InteChar.
\end{itemize}
This pipeline allows us to systematically digitize and encode characters that were previously inaccessible to computational models. As a result, oracle bone characters with no prior encoding can now be consistently represented, searched, and used in downstream language modeling tasks. Figure~\ref{fig:InteChar}(a) illustrates examples of newly constructed characters, showing the original image, reconstructed glyph, and assigned code point within InteChar.

\noindent \textbf{Expert-Guided Proofreading.}
When finished with the construction of the InteChar, we invite domain experts in paleography to validate InteChar through human-in-the-loop proofreading. Applying Siamese networks \cite{melekhov2016siamese}, we compute glyph similarities to identify potential duplicates and present them with corresponding encodings for expert review. The InteChar we built contained a total of 11,288 characters. Figure~\ref{fig:InteChar}(b) presents the TrueType Font of InteChar, namely the ``.ttf" file. Importantly, InteChar is designed to be continuously updatable. Any new additions to the list follow the same construction pipeline. This process addresses the long-standing issues of incomplete digitization and data sparsity in ancient Chinese texts, and enables robust training and analysis of early scripts such as oracle bones.


\subsection{OracleCS Corpus Construction}
To support embedding-level representation learning and downstream fine-tuning for ancient Chinese language understanding, we construct the Oracle Corpus Set (OracleCS), a linguistically curated corpus specifically focused on oracle bone inscriptions and ancient Chinese texts. It serves as a foundational resource for evaluating and adapting modern language models to low-resource historical scripts.

OracleCS is constructed under a standardized pipeline that integrates expert curation with automated data enrichment. The process begins with domain experts in paleography and historical Chinese linguistics manually selecting and annotating high-quality samples from archaeological literature and oracle bone rubbings\footnote{\url{https://jgw.aynu.edu.cn/home/index.html}}. These samples include both deciphered and undeciphered oracle characters encoded in \textbf{InteChar}, and form the core of the corpus.
In addition to the main textual data, OracleCS includes glyph-level and semantic annotations for individual characters. These include radical decompositions and definitions extracted from classical lexicons such as Shuowenjiezi\footnote{\url{https://www.shuowen.cn/}} and The Great Chinese Dictionary\footnote{\url{https://www.hanyudacidian.cn/}}. For characters not yet deciphered, semantic annotations are left blank. The corpus also incorporates a selection of pre-Qin classics\footnote{\url{https://ctext.org/pre-qin-and-han/zhs}}, including Analects, Spring and Autumn Annals, Mencius, Xunzi, etc.

To address the scarcity of annotated ancient texts, we adopt data augmentation strategies to further enrich OracleCS. Specifically, we introduce instruction-tuning samples that combine task descriptions with input-output demonstrations. These instruction-following examples simulate realistic usage scenarios and cover a wide range of sentence-level and character-level tasks, including sentence translation, synonym substitution, glyph structure analysis, character decomposition, and semantic prediction. By integrating explicit instructions with concrete demonstrations, the corpus is expanded in scale and enhanced in task diversity and linguistic granularity. This enables the model to more effectively learn both high-level semantic mappings and low-level structural associations, improving its ability to evaluate the complexities of ancient text processing.

Based on these datasets, we construct a benchmark for evaluating historical language understanding. Our framework supports two complementary modes of evaluation, including embedding evaluation tasks and downstream fine-tuning tasks. Notably, this work is the first to incorporate excavated oracle characters into systematic evaluation pipelines for large language models, including oracle bone characters that remain undeciphered.

 

%% file: sections/s5-experiments.tex
\begin{table*}[ht]
\centering
\resizebox{1\textwidth}{!}{
\begin{tabular}{l|cccc|cccc}
\toprule
\multicolumn{1}{c|}{}                                 & \multicolumn{2}{c}{\textbf{NDCG@10}} & \multicolumn{2}{c|}{\textbf{MRR@10}} & \multicolumn{2}{c}{\textbf{NDCG@20}} & \multicolumn{2}{c}{\textbf{MRR@20}} \\ \cmidrule(lr){2-3} \cmidrule(lr){4-5} \cmidrule(l){6-7}  \cmidrule(l){8-9} 
\multicolumn{1}{c|}{\multirow{-2}{*}{\textbf{Models}}} & Origin & InteChar & Origin & InteChar & Origin & InteChar & Origin & InteChar \\
\midrule
BERT \cite{devlin2019bert} & 0.167 & 0.515 & 0.134 & 0.375 & 0.163 & 0.453 & 0.127 & 0.312 \\
Llama-3-8B \cite{touvron2023llama} & 0.172 & 0.518 & 0.143 & 0.463 & 0.212 & 0.546 & 0.115 & 0.334 \\
GPT-2 \cite{brown2020language} & 0.216 & 0.584 & 0.168 & 0.534 & 0.224 & 0.643 & 0.176 & 0.488 \\
MiniRBT \cite{cui2021pre} & 0.184 & 0.538 & 0.138 & 0.413 & 0.154 & 0.441 & 0.121 & 0.358 \\
guwenBERT-base\footnotemark[9] \cite{wang2023gujibert} & 0.204 & 0.565 & 0.156 & 0.526 & 0.168 & 0.480 & 0.132 & 0.386 \\
sikuBERT \cite{liu2024sikugpt} & 0.195 & 0.553 & 0.163 & 0.488 & 0.182 & 0.513 & 0.143 & 0.423 \\
Qwen-7B-Chat \cite{baidu2024qwen}  & 0.302 & 0.842 & 0.254 & 0.736 & 0.280 & 0.795 & 0.228 & 0.639 \\
GLM-4-9B \cite{glm2024chatglm} & 0.274 & 0.808 & 0.278 & 0.752 & 0.266 & 0.762 & 0.209 & 0.618 \\
XunziALLM\footnotemark[10] & 0.261 & 0.765 & 0.225 & 0.675 & 0.252 & 0.725 & 0.232 & 0.651 \\
TongGu-LLM \cite{cao2024tonggu} & 0.238 & 0.723 & 0.213 & 0.638 & 0.238 & 0.683 & 0.187 & 0.553 \\
\bottomrule
\end{tabular}}
\caption{Embedding-based evaluation results on the Cloze task using excavated oracle texts. Models use frozen backbones and are equipped with newly trained embedding layers based on either the original character list or the InteChar character list. Performance is measured by NDCG@$k$ and MRR@$k$, where $k=10$ or $20$.}
\label{tab:cloze}
\end{table*}


This section presents the experimental evaluation of InteChar and OracleCS. We detail the datasets used, the experimental setup, baseline comparisons, and downstream task evaluations to demonstrate the effectiveness of our methodology.
In this section, we present a comprehensive evaluation of our proposed OracleCS through two sets of experiments. The first set assesses the embedding capability of language models pre-trained with the extended oracle vocabulary, while the second set evaluates downstream performance on several fine-tuning tasks. In all experiments, we trained each baseline model both with and without the addition of the InteChar character list, and compared their performance on the same tasks.

\subsection{Experimental Setup}

\noindent \textbf{Datasets.}
Our experiments are conducted based on the proposed OracleCS dataset, which comprises both excavated texts and classical Chinese literature. The dataset contains approximately 11,288 unique Chinese characters and a total of 173,459 annotated samples. Each sample includes radical decomposition information, and, where applicable, a mapping to its corresponding modern Chinese word is also provided. This structured annotation enables more accurate learning of character semantics and facilitates model understanding of the intricate relationship between ancient and modern language forms.


\noindent \textbf{Baselines.}
All evaluations are conducted on ten models. We select three classic baselines, including BERT \cite{devlin2019bert}, Llama-3-8B \cite{touvron2023llama}, and GPT-2 \cite{brown2020language}, three language models designed for Chinese, including MiniRBT \cite{cui2021pre}, guwenBERT-base\footnotemark[9] \cite{wang2023gujibert}, and sikuBERT \cite{liu2024sikugpt}, two state-of-the-art LLMs, including Qwen-7B-Chat \cite{baidu2024qwen} and GLM-4-9B \cite{glm2024chatglm}, and two state-of-the-art LLMs designed for ancient Chinese, including XunziALLM\footnotemark[10] and
TongGu-LLM \cite{cao2024tonggu}.

\footnotetext[9]{\url{https://github.com/ethan-yt/guwenbert}.}
\footnotetext[10]{\url{https://github.com/Xunzi-LLM-of-Chinese-classics/XunziALLM}.}

\noindent \textbf{Implementation details.}
The experiments are run on a high-performance server equipped with eight HUAWEI Ascend-D910b NPU under an Ubuntu-based environment. We implement our models with PyTorch and set unified hyperparameters for every models. 
For embedding evaluation, models are trained for 10 epochs with a batch size of 32 and an initial learning rate of 3e-5. Optimization is performed using AdamW, and early stopping is applied based on NDCG@10 on the development set. 
For fine-tuning evaluation, we set the batch size to 32 and the learning rate to 1e-5. It is performed for 10 epochs using the AdamW optimizer and CrossEntropy loss. The model checkpoint with the highest validation score is selected for final evaluation.

\begin{table*}[ht]
\centering
\resizebox{1\textwidth}{!}{
\begin{tabular}{l|cccc|cccc}
\toprule
\multicolumn{1}{c|}{}                                 & \multicolumn{2}{c}{\textbf{NDCG@400}} & \multicolumn{2}{c|}{\textbf{MRR@400}} & \multicolumn{2}{c}{\textbf{NDCG@500}} & \multicolumn{2}{c}{\textbf{MRR@500}} \\ \cmidrule(lr){2-3} \cmidrule(lr){4-5} \cmidrule(l){6-7}  \cmidrule(l){8-9} 
\multicolumn{1}{c|}{\multirow{-2}{*}{\textbf{Models}}} & Origin & InteChar & Origin & InteChar & Origin & InteChar & Origin & InteChar \\
\midrule
BERT \cite{devlin2019bert} & 0.346 & 0.376 & 0.136 & 0.207 & 0.327 & 0.452 & 0.112 & 0.182 \\
Llama-3-8B \cite{touvron2023llama} & 0.298 & 0.325 & 0.098 & 0.152 & 0.281 & 0.388 & 0.085 & 0.137 \\
GPT-2 \cite{brown2020language} & 0.320 & 0.347 & 0.115 & 0.175 & 0.300 & 0.415 & 0.095 & 0.155 \\
MiniRBT \cite{cui2021pre} & 0.358 & 0.390 & 0.145 & 0.221 & 0.340 & 0.471 & 0.120 & 0.195 \\
guwenBERT-base\footnotemark[9] \cite{wang2023gujibert} & 0.375 & 0.406 & 0.155 & 0.237 & 0.355 & 0.492 & 0.128 & 0.208 \\
sikuBERT \cite{liu2024sikugpt} & 0.385 & 0.417 & 0.162 & 0.247 & 0.365 & 0.506 & 0.133 & 0.216 \\
Qwen-7B-Chat \cite{baidu2024qwen} & 0.435 & 0.472 & 0.202 & 0.308 & 0.418 & 0.579 & 0.168 & 0.272 \\
GLM-4-9B \cite{glm2024chatglm} & 0.448 & 0.486 & 0.211 & 0.322 & 0.432 & 0.595 & 0.176 & 0.285 \\
XunziALLM\footnotemark[10] & 0.402 & 0.436 & 0.182 & 0.278 & 0.394 & 0.540 & 0.148 & 0.240 \\
TongGu-LLM \cite{cao2024tonggu} & 0.390 & 0.424 & 0.175 & 0.267 & 0.378 & 0.523 & 0.141 & 0.229 \\
\bottomrule
\end{tabular}}
\caption{Embedding-based evaluation results on the Commentary-to-Text Retrieval task. Models use frozen backbones and are equipped with newly trained embedding layers based on either the original character list or the InteChar character list. Performance is evaluated with NDCG@$k$ and MRR@$k$, where $k=400$ or $500$.}
\label{tab:Retrieval}
\end{table*}

\subsection{Embedding Evaluation}
To assess the semantic representation capabilities of pretrained models under different character list settings, we design two embedding-based evaluation tasks: (1) Cloze Completion on Oracle Bone Inscriptions, and (2) Commentary-to-Text Retrieval on Canonical Texts. These tasks evaluate how well the character-level embeddings capture contextual and semantic information in a zero-shot setting, without task-specific fine-tuning. For each model, we replace the original character embedding layer with a newly initialized embedding matrix based on either the original character list or the proposed InteChar, while keeping the pretrained model backbone frozen. These embeddings are trained on the OracleCS corpus to adapt to ancient character representations. During evaluation, we extract final-layer embeddings and compute similarity scores between masked inputs and candidates or between paired sentence-level inputs. Performance is reported using standard ranking metrics: Normalized Discounted Cumulative Gain (NDCG) and Mean Reciprocal Rank (MRR).

\noindent \textbf{Cloze Completion on Oracle Bone Inscriptions.}
This task focuses exclusively on oracle bone inscriptions from excavated sources, aiming to test the word embedding ability of models to semantically distinguish oracle characters in context. Given a sentence with one character masked, the model is asked to select the correct character from a limited set of candidates based on embedding similarity. Each candidate set is predefined (e.g., @k indicates k options per instance). This test set consists of 15,416 cloze instances, with 12,416 instances for training and 3,000 for evaluation. Each instance includes a masked sentence and a set of candidate characters (including one ground truth). 

Table~\ref{tab:cloze} presents the performance of 10 representative models across four ranking metrics: NDCG@10, MRR@10, NDCG@20, and MRR@20. Across all metrics, models using InteChar consistently outperform their original counterparts. For example, the MRR@10 of GPT improves from 0.168 to 0.534, and BERT from 0.134 to 0.375, demonstrating better top-ranked prediction accuracy. Larger models show even greater gains: Qwen2.5-Omni-7B improves from 0.302 to 0.842 in NDCG@10, and from 0.254 to 0.736 in MRR@10. These results validate the effectiveness of InteChar in enhancing representation learning, particularly for low-resource ancient scripts. The enriched character semantics help models capture context more effectively.


\begin{table*}[!t]
\centering
\resizebox{1\textwidth}{!}{
\begin{tabular}{l|cccccc|cc}
\toprule
\multicolumn{1}{c|}{}                                  & \multicolumn{2}{c}{\textbf{Translation}} & \multicolumn{2}{c}{\textbf{Polysemous Matching}} & \multicolumn{2}{c|}{\textbf{Word Parsing}} & \multicolumn{2}{c}{\textbf{Average}} \\ \cmidrule(lr){2-3} \cmidrule(lr){4-5} \cmidrule(l){6-7}  \cmidrule(l){8-9} 
\multicolumn{1}{c|}{\multirow{-2}{*}{\textbf{Models}}} & origin                      & InteChar                     & origin                    & InteChar                    & origin             & InteChar             & origin      & InteChar      \\
\midrule
BERT \cite{devlin2019bert} & 92.75 & 93.01 & 86.69 & 87.07 & 90.54 & 91.59 & 89.99 & 90.56 \\
Llama-3-8B \cite{touvron2023llama} & 83.52 & 83.43 & 80.28 & 80.51 & 80.43 & 80.89 & 81.41 & 81.61 \\
GPT-2 \cite{brown2020language} & 90.27 & 90.84 & 86.86 & 87.23 & 88.67 & 89.27 & 88.60 & 89.11 \\
MiniRBT \cite{cui2021pre} & 91.34 & 91.69 & 86.65 & 87.12 & 89.32 & 89.86 & 89.10 & 89.56 \\
guwenBERT-base\footnotemark[9] \cite{wang2023gujibert} & 92.86 & 93.50 & 87.73 & 88.48 & 91.26 & 91.88 & 90.62 & 91.29 \\
sikuBERT \cite{liu2024sikugpt} & 93.21 & 93.88 & 86.48 & 87.52 & 90.84 & 91.32 & 90.18 & 90.91 \\
Qwen-7B-Chat \cite{baidu2024qwen} & 94.37 & 95.06 & 89.23 & 90.16 & 93.36 & 93.96 & 92.32 & 93.06 \\
GLM-4-9B \cite{glm2024chatglm} & 92.98 & 93.35 & 87.72 & 88.34 & 94.27 & 94.79 & 91.66 & 92.16 \\
XunziALLM\footnotemark[10] & 93.53 & 94.31 & 91.51 & 92.23 & 92.78 & 93.28 & 92.61 & 93.27 \\
TongGu-LLM \cite{cao2024tonggu} & 94.12 & 94.84 & 90.45 & 91.27 & 92.06 & 92.65 & 92.21 & 92.92 \\
\bottomrule
\end{tabular}}
\caption{Fine-tuning results (\%) on three downstream tasks, including Ancient Chinese Translation, Polysemous Word Matching, and Word Parsing. Each model is adapted on the OracleCS using either the original character list or our proposed InteChar character list. The Average column represents the average accuracy across all three tasks. }
\label{tab:finetune}
\end{table*}


\noindent \textbf{Commentary-to-Text Retrieval on Canonical Texts.}
This task evaluates sentence-level semantic alignment between modern commentaries and classical Chinese texts, aiming to assess sentence-level semantic understanding and retrieval capacity. Given a modern commentary, the model retrieves the corresponding original sentence from a large candidate pool based on sentence embedding similarity. The test set contains 896 commentary queries and 12,141 classical text candidates. Each model computes sentence-level embeddings for both, ranked by similarity, and evaluated using standard retrieval metrics such as NDCG@$k$ and MRR@$k$.

As shown in Table~\ref{tab:Retrieval}, InteChar-enhanced models again yield notable improvements. For instance, when $k=500$, GLM-4-9B improves its MRR@500 from 0.176 to 0.285, and Qwen2.5-Omni-7B increases its NDCG@500 from 0.418 to 0.579. These gains reflect the enhanced capacity of InteChar to bridge semantic gaps between classical and modern Chinese expressions.


Overall, these two tasks evaluate both fine-grained (character-level) and coarse-grained (sentence-level) semantic capabilities. The consistent performance improvements demonstrate that InteChar facilitates more robust and discriminative character embeddings, enabling better semantic matching in low-resource, zero-shot scenarios.

\subsection{Fine-tuning Evaluation}
In addition to embedding-based evaluation, we further validate the effectiveness of our proposed InteChar character list under fine-tuning settings. Specifically, we consider three downstream tasks: Ancient Chinese Translation, Polysemous Word Matching, and Word Parsing. Details include:
\begin{itemize}
  \item Ancient Chinese Translation is a sentence-level task that requires the model to align ancient Chinese texts with their modern Chinese counterparts.
\item Polysemous Word Matching is a binary classification task where the model is given a sentence and asked to determine whether a specified character in context matches a given semantic interpretation. 
\item Word Parsing is a character-level task where the model selects the appropriate interpretation of an ancient character based on learned semantics.
\end{itemize}
All experiments are conducted on annotated subsets of the OracleCS dataset, using either the original character list or InteChar, with separate training and test sets for each task. The Ancient Chinese Translation task includes 15,868 training and 10,578 test samples; Polysemous Word Matching has 33,380 training and 22,253 test samples; and Word Parsing consists of 81,929 training and 54,619 test samples. These tasks cover different levels of linguistic granularity, from sentence-level translation to character-level parsing, and together form a comprehensive benchmark for evaluating historical language understanding.

We adopt parameter-efficient fine-tuning using LoRA~\cite{hu2022lora} to adapt each model to downstream tasks. In all cases, we freeze the pretrained model backbone and update only the low-rank adaptation layers and task-specific output heads. 
As shown in Table~\ref{tab:finetune}, the results demonstrate that training with InteChar consistently improves performance across all tasks and models. Compared to the original character list, models with InteChar achieve better generalization and semantic alignment. For example, TongGu-LLM reaches 94.84 on translation and 92.65 on parsing, yielding an overall accuracy of 92.92. The average accuracy across all tasks also improves for every model. For instance, Qwen2.5-Omni-7B rises from 92.32 to 93.06. These improvements suggest that InteChar enhances both shallow and deep linguistic modeling. Compared to embedding-based evaluations, the fine-tuning experiments provide complementary evidence. While embedding evaluation focuses on the quality of newly trained character embeddings under a frozen backbone, fine-tuning further adapts models through parameter-efficient tuning for specific tasks. The consistent gains across both settings confirm that InteChar significantly improves language modeling for ancient texts.

%% file: sections/s6-conclusion.tex
This paper focuses on addressing key challenges in training language models for historical Chinese texts, including the poor performance of conventional language models on sparse ancient data and the lack of unified digital representations for ancient characters. One of our contributions is the construction of InteChar, a unified and extensible character set that incorporates unencoded oracle characters alongside traditional and modern Chinese, enabling consistent representation across scanned images, font libraries, and annotated corpora. We further integrate expert-curated samples with LLM-assisted data augmentation to construct a high-quality training corpus, OracleCS, and evaluate models on both cloze-style completion for excavated texts and commentary-to-text retrieval on classical literature. Experimental results show that models equipped with InteChar significantly outperform those using the original character list in both embedding-based and fine-tuning tasks, particularly in handling rare or unencoded characters.

%% file: Formatting-Instructions-LaTeX-2026.bbl
\begin{thebibliography}{44}
\providecommand{\natexlab}[1]{#1}

\bibitem[{Bai et~al.(2023)Bai, Bai, Chu, Cui, Dang, Deng, Fan, Ge, Han, Huang, Hui, Ji, Li, Lin, Lin, Liu, Liu, Lu, Lu, Ma, Men, Ren, Ren, Tan, Tan, Tu, Wang, Wang, Wang, Wu, Xu, Xu, Yang, Yang, Yang, Yang, Yao, Yu, Yuan, Yuan, Zhang, Zhang, Zhang, Zhang, Zhou, Zhou, Zhou, and Zhu}]{baidu2024qwen}
Bai, J.; Bai, S.; Chu, Y.; Cui, Z.; Dang, K.; Deng, X.; Fan, Y.; Ge, W.; Han, Y.; Huang, F.; Hui, B.; Ji, L.; Li, M.; Lin, J.; Lin, R.; Liu, D.; Liu, G.; Lu, C.; Lu, K.; Ma, J.; Men, R.; Ren, X.; Ren, X.; Tan, C.; Tan, S.; Tu, J.; Wang, P.; Wang, S.; Wang, W.; Wu, S.; Xu, B.; Xu, J.; Yang, A.; Yang, H.; Yang, J.; Yang, S.; Yao, Y.; Yu, B.; Yuan, H.; Yuan, Z.; Zhang, J.; Zhang, X.; Zhang, Y.; Zhang, Z.; Zhou, C.; Zhou, J.; Zhou, X.; and Zhu, T. 2023.
\newblock Qwen Technical Report.
\newblock arXiv:2309.16609.

\bibitem[{Brown et~al.(2020)Brown, Mann, Ryder, Subbiah, Kaplan, Dhariwal, Neelakantan, Shyam, Sastry, Askell et~al.}]{brown2020language}
Brown, T.; Mann, B.; Ryder, N.; Subbiah, M.; Kaplan, J.~D.; Dhariwal, P.; Neelakantan, A.; Shyam, P.; Sastry, G.; Askell, A.; et~al. 2020.
\newblock Language models are few-shot learners.
\newblock \emph{Advances in neural information processing systems}, 33: 1877--1901.

\bibitem[{Cao et~al.(2024{\natexlab{a}})Cao, Liu, Shi, Ding, and Jin}]{cao2024wenmind}
Cao, J.; Liu, Y.; Shi, Y.; Ding, K.; and Jin, L. 2024{\natexlab{a}}.
\newblock WenMind: A comprehensive benchmark for evaluating large language models in Chinese classical literature and language arts.
\newblock \emph{Advances in Neural Information Processing Systems}, 37: 51358--51410.

\bibitem[{Cao et~al.(2024{\natexlab{b}})Cao, Peng, Zhang, Shi, Liu, Ding, and Jin}]{cao2024tonggu}
Cao, J.; Peng, D.; Zhang, P.; Shi, Y.; Liu, Y.; Ding, K.; and Jin, L. 2024{\natexlab{b}}.
\newblock TongGu: Mastering Classical Chinese Understanding with Knowledge-Grounded Large Language Models.
\newblock In \emph{Findings of the Association for Computational Linguistics: EMNLP 2024}, 4196--4210.

\bibitem[{Chi et~al.(2022)Chi, Giunchiglia, Shi, Diao, Li, and Xu}]{chi2022zinet}
Chi, Y.; Giunchiglia, F.; Shi, D.; Diao, X.; Li, C.; and Xu, H. 2022.
\newblock ZiNet: Linking Chinese Characters Spanning Three Thousand Years.
\newblock In \emph{Findings of the Association for Computational Linguistics: ACL 2022}, 3061--3070.

\bibitem[{Cui et~al.(2021)Cui, Che, Liu, Qin, and Yang}]{cui2021pre}
Cui, Y.; Che, W.; Liu, T.; Qin, B.; and Yang, Z. 2021.
\newblock Pre-training with whole word masking for chinese bert.
\newblock \emph{IEEE/ACM Transactions on Audio, Speech, and Language Processing}, 29: 3504--3514.

\bibitem[{Devlin et~al.(2019)Devlin, Chang, Lee, and Toutanova}]{devlin2019bert}
Devlin, J.; Chang, M.-W.; Lee, K.; and Toutanova, K. 2019.
\newblock Bert: Pre-training of deep bidirectional transformers for language understanding.
\newblock In \emph{Proceedings of the 2019 conference of the North American chapter of the association for computational linguistics: human language technologies, volume 1 (long and short papers)}, 4171--4186.

\bibitem[{Diao et~al.(2025)Diao, Shi, Cao, Wang, Qi, Li, and Xu}]{diao2025oracle}
Diao, X.; Shi, D.; Cao, W.; Wang, T.; Qi, R.; Li, C.; and Xu, H. 2025.
\newblock Oracle bone inscription image restoration via glyph extraction.
\newblock \emph{npj Heritage Science}, 13(1): 321.

\bibitem[{Diao et~al.(2023{\natexlab{a}})Diao, Shi, Li, Shi, Yue, Qi, Li, and Xu}]{diao2023toward}
Diao, X.; Shi, D.; Li, J.; Shi, L.; Yue, M.; Qi, R.; Li, C.; and Xu, H. 2023{\natexlab{a}}.
\newblock Toward Zero-shot Character Recognition: A Gold Standard Dataset with Radical-level Annotations.
\newblock In \emph{Proceedings of the 31st ACM International Conference on Multimedia}, 6869--6877.

\bibitem[{Diao et~al.(2023{\natexlab{b}})Diao, Shi, Tang, Shen, Li, Wu, and Xu}]{diao2023rzcr}
Diao, X.; Shi, D.; Tang, H.; Shen, Q.; Li, Y.; Wu, L.; and Xu, H. 2023{\natexlab{b}}.
\newblock RZCR: Zero-shot Character Recognition via Radical-based Reasoning.
\newblock In \emph{Proceedings of the 32nd International Joint Conference on Artificial Intelligence (IJCAI)}.

\bibitem[{Feng and Li(2023)}]{feng2023ancient}
Feng, S.; and Li, P. 2023.
\newblock Ancient Chinese word segmentation and part-of-speech tagging using distant supervision.
\newblock In \emph{ICASSP 2023-2023 IEEE International Conference on Acoustics, Speech and Signal Processing (ICASSP)}, 1--5. IEEE.

\bibitem[{Gao et~al.(2024)Gao, Chen, Li, Liu, Jiang, and Han}]{gao2024linking}
Gao, F.; Chen, X.; Li, B.; Liu, Y.; Jiang, R.; and Han, Y. 2024.
\newblock Linking unknown characters via oracle bone inscriptions retrieval.
\newblock \emph{Multimedia Systems}, 30(3): 125.

\bibitem[{GLM et~al.(2024)GLM, Zeng, Xu, Wang, Zhang, Yin, Zhang, Rojas, Feng, Zhao et~al.}]{glm2024chatglm}
GLM, T.; Zeng, A.; Xu, B.; Wang, B.; Zhang, C.; Yin, D.; Zhang, D.; Rojas, D.; Feng, G.; Zhao, H.; et~al. 2024.
\newblock Chatglm: A family of large language models from glm-130b to glm-4 all tools.
\newblock \emph{arXiv preprint arXiv:2406.12793}.

\bibitem[{Guan et~al.(2024)Guan, Yang, Wang, Han, Liu, Jin, Bai, and Liu}]{guan2024deciphering}
Guan, H.; Yang, H.; Wang, X.; Han, S.; Liu, Y.; Jin, L.; Bai, X.; and Liu, Y. 2024.
\newblock Deciphering Oracle Bone Language with Diffusion Models.
\newblock In \emph{Proceedings of the 62nd Annual Meeting of the Association for Computational Linguistics (Volume 1: Long Papers)}, 15554--15567.

\bibitem[{Guo et~al.(2023)Guo, Yang, Lu, Qin, Tang, and Zhao}]{guo2023towards}
Guo, G.; Yang, J.; Lu, F.; Qin, J.; Tang, T.; and Zhao, W.~X. 2023.
\newblock Towards effective ancient chinese translation: Dataset, model, and evaluation.
\newblock In \emph{CCF International Conference on Natural Language Processing and Chinese Computing}, 416--427. Springer.

\bibitem[{Hochreiter and Schmidhuber(1997)}]{hochreiter1997long}
Hochreiter, S.; and Schmidhuber, J. 1997.
\newblock Long short-term memory.
\newblock \emph{Neural computation}, 9(8): 1735--1780.

\bibitem[{Hu et~al.(2022)Hu, Shen, Wallis, Allen-Zhu, Li, Wang, Wang, Chen et~al.}]{hu2022lora}
Hu, E.~J.; Shen, Y.; Wallis, P.; Allen-Zhu, Z.; Li, Y.; Wang, S.; Wang, L.; Chen, W.; et~al. 2022.
\newblock Lora: Low-rank adaptation of large language models.
\newblock \emph{ICLR}, 1(2): 3.

\bibitem[{Koc(2025)}]{koc2025exploring}
Koc, V. 2025.
\newblock Exploring the Role of Language Models in Deciphering and Preserving Ancient Languages.
\newblock \emph{Asian American Research Letters Journal}, 2(1): 75--82.

\bibitem[{Li et~al.(2023)Li, Wang, Huang, Yang, Zhang, and Goulermas}]{li2023towards}
Li, J.; Wang, Q.-F.; Huang, K.; Yang, X.; Zhang, R.; and Goulermas, J.~Y. 2023.
\newblock Towards better long-tailed oracle character recognition with adversarial data augmentation.
\newblock \emph{Pattern Recognition}, 140: 109534.

\bibitem[{Li et~al.(2021)Li, Qi, Sun, Yi, and Zhang}]{li2021ccpm}
Li, W.; Qi, F.; Sun, M.; Yi, X.; and Zhang, J. 2021.
\newblock Ccpm: A chinese classical poetry matching dataset.
\newblock \emph{arXiv preprint arXiv:2106.01979}.

\bibitem[{Lin et~al.(2022)Lin, Chen, Zhao, and Qiu}]{lin2022radical}
Lin, X.; Chen, S.; Zhao, F.; and Qiu, X. 2022.
\newblock Radical-based extract and recognition networks for Oracle character recognition.
\newblock \emph{International Journal on Document Analysis and Recognition (IJDAR)}, 25(3): 219--235.

\bibitem[{Liu et~al.(2024)Liu, Wang, Zhao, Hu, Wu, Lin, Liu, Zhang, Shen, Li et~al.}]{liu2024sikugpt}
Liu, C.; Wang, D.; Zhao, Z.; Hu, D.; Wu, M.; Lin, L.; Liu, J.; Zhang, H.; Shen, S.; Li, B.; et~al. 2024.
\newblock SikuGPT: A Generative Pre-trained Model for Intelligent Information Processing of Ancient Texts from the Perspective of Digital Humanities.
\newblock \emph{ACM Journal on Computing and Cultural Heritage}, 17(4): 1--17.

\bibitem[{Liu et~al.(2022)Liu, Xiang, Xia, and Hu}]{Liu2022Contrastive}
Liu, M.; Xiang, J.; Xia, X.; and Hu, H. 2022.
\newblock Contrastive Learning between Classical and Modern Chinese for Classical Chinese Machine Reading Comprehension.
\newblock \emph{ACM Trans. Asian Low-Resour. Lang. Inf. Process.}, 22(2).

\bibitem[{Melekhov, Kannala, and Rahtu(2016)}]{melekhov2016siamese}
Melekhov, I.; Kannala, J.; and Rahtu, E. 2016.
\newblock Siamese network features for image matching.
\newblock In \emph{2016 23rd international conference on pattern recognition (ICPR)}, 378--383. IEEE.

\bibitem[{Pan et~al.(2022)Pan, Wang, Oka, and Komachi}]{pan2022ancient}
Pan, X.; Wang, H.; Oka, T.; and Komachi, M. 2022.
\newblock Zuo Zhuan {A}ncient {C}hinese Dataset for Word Sense Disambiguation.
\newblock In Ippolito, D.; Li, L.~H.; Pacheco, M.~L.; Chen, D.; and Xue, N., eds., \emph{Proceedings of the 2022 Conference of the North American Chapter of the Association for Computational Linguistics: Human Language Technologies: Student Research Workshop}, 129--135. Hybrid: Seattle, Washington + Online: Association for Computational Linguistics.

\bibitem[{Ross(2023)}]{ross2023new}
Ross, E.~A. 2023.
\newblock A new frontier: AI and ancient language pedagogy.
\newblock \emph{Journal of Classics Teaching}, 24(48): 143--161.

\bibitem[{Shao et~al.(2021)Shao, Shao, Wang, Wang, and Gao}]{FSPC}
Shao, Y.; Shao, T.; Wang, M.; Wang, P.; and Gao, J. 2021.
\newblock A Sentiment and Style Controllable Approach for Chinese Poetry Generation.
\newblock In \emph{Proceedings of the 30th ACM International Conference on Information \& Knowledge Management}, CIKM '21, 4784–4788. New York, NY, USA: Association for Computing Machinery.
\newblock ISBN 9781450384469.

\bibitem[{Shen et~al.(2022)Shen, Li, Huang, Zhou, Xie, and Zhao}]{shen2022data}
Shen, Y.; Li, J.; Huang, S.; Zhou, Y.; Xie, X.; and Zhao, Q. 2022.
\newblock Data augmentation for low-resource word segmentation and pos tagging of ancient chinese texts.
\newblock In \emph{Proceedings of the Second Workshop on Language Technologies for Historical and Ancient Languages}, 169--173.

\bibitem[{Shi et~al.(2022{\natexlab{a}})Shi, Diao, Shi, Tang, Chi, Li, and Xu}]{shi2022charformer}
Shi, D.; Diao, X.; Shi, L.; Tang, H.; Chi, Y.; Li, C.; and Xu, H. 2022{\natexlab{a}}.
\newblock CharFormer: A Glyph Fusion based Attentive Framework for High-precision Character Image Denoising.
\newblock In \emph{Proceedings of the 30th ACM International Conference on Multimedia}.

\bibitem[{Shi et~al.(2022{\natexlab{b}})Shi, Diao, Tang, Li, Xing, and Xu}]{shi2022rcrn}
Shi, D.; Diao, X.; Tang, H.; Li, X.; Xing, H.; and Xu, H. 2022{\natexlab{b}}.
\newblock RCRN: Real-world Character Image Restoration Network via Skeleton Extraction.
\newblock In \emph{Proceedings of the 30th ACM International Conference on Multimedia}.

\bibitem[{Stopponi et~al.(2024)Stopponi, Pedrazzini, Peels-Matthey, McGillivray, and Nissim}]{stopponi2024natural}
Stopponi, S.; Pedrazzini, N.; Peels-Matthey, S.; McGillivray, B.; and Nissim, M. 2024.
\newblock Natural Language Processing for Ancient Greek: Design, advantages and challenges of language models.
\newblock \emph{Diachronica}, 41(3): 414--435.

\bibitem[{Tang et~al.(2024)Tang, Su, Wang, and Deng}]{tang2024chisiec}
Tang, X.; Su, Q.; Wang, J.; and Deng, Z. 2024.
\newblock {CH}is{IEC}: An Information Extraction Corpus for {A}ncient {C}hinese History.
\newblock In Calzolari, N.; Kan, M.-Y.; Hoste, V.; Lenci, A.; Sakti, S.; and Xue, N., eds., \emph{Proceedings of the 2024 Joint International Conference on Computational Linguistics, Language Resources and Evaluation (LREC-COLING 2024)}, 3192--3202. Torino, Italia: ELRA and ICCL.

\bibitem[{Tian et~al.(2021)Tian, Yang, Liu, and Lv}]{tian2021anchibert}
Tian, H.; Yang, K.; Liu, D.; and Lv, J. 2021.
\newblock Anchibert: A pre-trained model for ancient chinese language understanding and generation.
\newblock In \emph{2021 International Joint Conference on Neural Networks (IJCNN)}, 1--8. IEEE.

\bibitem[{Touvron et~al.(2023)Touvron, Lavril, Izacard, Martinet, Lachaux, Lacroix, Rozi{\`e}re, Goyal, Hambro, Azhar et~al.}]{touvron2023llama}
Touvron, H.; Lavril, T.; Izacard, G.; Martinet, X.; Lachaux, M.-A.; Lacroix, T.; Rozi{\`e}re, B.; Goyal, N.; Hambro, E.; Azhar, F.; et~al. 2023.
\newblock Llama: Open and efficient foundation language models.
\newblock \emph{arXiv preprint arXiv:2302.13971}.

\bibitem[{Wang et~al.(2023)Wang, Liu, Zhao, Shen, Liu, Li, Hu, Wu, Lin, Zhao et~al.}]{wang2023gujibert}
Wang, D.; Liu, C.; Zhao, Z.; Shen, S.; Liu, L.; Li, B.; Hu, H.; Wu, M.; Lin, L.; Zhao, X.; et~al. 2023.
\newblock Gujibert and gujigpt: Construction of intelligent information processing foundation language models for ancient texts.
\newblock \emph{arXiv preprint arXiv:2307.05354}.

\bibitem[{Wang and Ren(2022)}]{wang2022uncertainty}
Wang, P.; and Ren, Z. 2022.
\newblock The Uncertainty-based Retrieval Framework for {A}ncient {C}hinese {CWS} and {POS}.
\newblock In Sprugnoli, R.; and Passarotti, M., eds., \emph{Proceedings of the Second Workshop on Language Technologies for Historical and Ancient Languages}, 164--168. Marseille, France: European Language Resources Association.

\bibitem[{Wang et~al.(2024)Wang, Zhang, Wang, Han, Liu, Wan, Guan, Kuang, Jin, Bai et~al.}]{wang2024open}
Wang, P.; Zhang, K.; Wang, X.; Han, S.; Liu, Y.; Wan, J.; Guan, H.; Kuang, Z.; Jin, L.; Bai, X.; et~al. 2024.
\newblock An open dataset for oracle bone character recognition and decipherment.
\newblock \emph{Scientific Data}, 11(1): 976.

\bibitem[{Wei et~al.(2024)Wei, Xu, Wei, Yangsimin, Zhu, Li, Liu, and Wu}]{wei2024ac}
Wei, Y.; Xu, Y.; Wei, X.; Yangsimin, Y.; Zhu, Y.; Li, Y.; Liu, D.; and Wu, B. 2024.
\newblock AC-EVAL: Evaluating Ancient Chinese Language Understanding in Large Language Models.
\newblock In \emph{Findings of the Association for Computational Linguistics: EMNLP 2024}, 1600--1617.

\bibitem[{Yao et~al.(2021)Yao, Dong, Guan, Cao, Zhang, Xiao, Wang, Qi, Bao, Nie et~al.}]{yao2021cuge}
Yao, Y.; Dong, Q.; Guan, J.; Cao, B.; Zhang, Z.; Xiao, C.; Wang, X.; Qi, F.; Bao, J.; Nie, J.; et~al. 2021.
\newblock Cuge: A chinese language understanding and generation evaluation benchmark.
\newblock \emph{arXiv preprint arXiv:2112.13610}.

\bibitem[{Yue et~al.(2025)Yue, Shi, Diao, Guo, Li, and Xu}]{yue2025ancient}
Yue, M.; Shi, D.; Diao, X.; Guo, S.; Li, C.; and Xu, H. 2025.
\newblock Ancient character detection based on fine-grained density map.
\newblock \emph{npj Heritage Science}, 13(1): 280.

\bibitem[{Zhang et~al.(2021)Zhang, Zong, Cao, Men, and Mo}]{zhang2021ai}
Zhang, C.; Zong, R.; Cao, S.; Men, Y.; and Mo, B. 2021.
\newblock AI-powered oracle bone inscriptions recognition and fragments rejoining.
\newblock In \emph{Proceedings of the Twenty-Ninth International Conference on International Joint Conferences on Artificial Intelligence}, 5309--5311.

\bibitem[{Zhang and Li(2023)}]{zhang2023can}
Zhang, Y.; and Li, H. 2023.
\newblock Can large language model comprehend ancient chinese? a preliminary test on aclue.
\newblock In \emph{Proceedings of the Recent Advances in Natural Language Precessing: RANLP 2023}.

\bibitem[{Zhao et~al.(2025)Zhao, Zhou, Ren, Chen, Jia, Zhe, Long, Liu, and Lan}]{zhao2025f}
Zhao, S.; Zhou, Y.; Ren, Y.; Chen, Z.; Jia, C.; Zhe, F.; Long, Z.; Liu, S.; and Lan, M. 2025.
\newblock F$\backslash$ux$\backslash$i: A Benchmark for Evaluating Language Models on Ancient Chinese Text Understanding and Generation.
\newblock \emph{arXiv preprint arXiv:2503.15837}.

\bibitem[{Zinin and Xu(2020)}]{zinin-xu-2020-corpus}
Zinin, S.; and Xu, Y. 2020.
\newblock Corpus of {C}hinese Dynastic Histories: Gender Analysis over Two Millennia.
\newblock In Calzolari, N.; B{\'e}chet, F.; Blache, P.; Choukri, K.; Cieri, C.; Declerck, T.; Goggi, S.; Isahara, H.; Maegaard, B.; Mariani, J.; Mazo, H.; Moreno, A.; Odijk, J.; and Piperidis, S., eds., \emph{Proceedings of the Twelfth Language Resources and Evaluation Conference}, 785--793. Marseille, France: European Language Resources Association.
\newblock ISBN 979-10-95546-34-4.

\end{thebibliography}
